\documentclass[journal]{IEEEtran}
\IEEEoverridecommandlockouts

\usepackage{cite}
\usepackage{amsmath,amssymb,amsfonts,amsthm} %
\usepackage{graphicx}
\usepackage{textcomp}
\usepackage{xcolor}
\usepackage{algorithmic} %
\usepackage{url}
\usepackage{microtype} %
\usepackage{booktabs}  %

\usepackage[printonlyused,withpage,nolist,nohyperlinks]{acronym}
\usepackage{xspace}
\usepackage{multirow}
\usepackage{diagbox}
\usepackage{subfig}

\begin{document}

\title{Distributed AI Agents for Cognitive Underwater Robot Autonomy
}

\author{Markus~Buchholz, %
        Ignacio~Carlucho, %
        Michele~Grimaldi, %
        and~Yvan~R.~Petillot %
\thanks{M. Buchholz, I. Carlucho, M. Grimaldi, and Y.R. Petillot are with the School of Engineering \& Physical Sciences, Heriot-Watt University, Edinburgh, UK. e-mail: m.buchholz@hw.ac.uk.}} %

\maketitle

\begin{abstract} Achieving robust cognitive autonomy in robots navigating complex, unpredictable environments remains a fundamental challenge in robotics. This paper presents Underwater Robot Self-Organizing Autonomy (UROSA), a groundbreaking architecture leveraging distributed Large Language Model AI agents integrated within the Robot Operating System 2 (ROS 2) framework to enable advanced cognitive capabilities in Autonomous Underwater Vehicles. UROSA decentralises cognition into specialised AI agents responsible for multimodal perception, adaptive reasoning, dynamic mission planning, and real-time decision-making. Central innovations include flexible agents dynamically adapting their roles, retrieval-augmented generation utilising vector databases for efficient knowledge management, reinforcement learning-driven behavioural optimisation, and autonomous on-the-fly ROS 2 node generation for runtime functional extensibility. Extensive empirical validation demonstrates UROSA's promising adaptability and reliability through realistic underwater missions in simulation and real-world deployments, showing significant advantages over traditional rule-based architectures in handling unforeseen scenarios, environmental uncertainties, and novel mission objectives. This work not only advances underwater autonomy but also establishes a scalable, safe, and versatile cognitive robotics framework capable of generalising to a diverse array of real-world applications.\footnote{UROSA project webpage: \url{https://markusbuchholz.github.io/urosa.html}} \end{abstract}

\section{Introduction}

The pursuit of truly autonomous robots capable of effectively navigating and interacting within complex, unstructured environments remains a grand challenge in the field of robotics \cite{RussellNorvig2020}. For decades, the dominant paradigm has been rooted in traditional robotic systems, which rely on modular architectures and meticulously pre-defined, rule-based algorithms \cite{Nilsson1984}. While these systems have demonstrated notable success in controlled and predictable settings such as factory automation, they increasingly reveal their limitations when confronted with the inherent dynamism and uncertainty of real-world scenarios \cite{Brooks1991}.

Legacy approaches often struggle with novel or unforeseen situations, demanding extensive manual reprogramming for even minor environmental changes and fundamentally lacking the adaptability needed for open-ended tasks \cite{Arkin1998}. This inherent inflexibility prevents them from truly operating in a \textit{self-playing} manner, where the system autonomously adapts and solves problems without continuous human intervention.

Within this evolving landscape, the rapid advancement of Large Language Models (LLMs) and Vision Language Models (VLMs) offers a compelling pathway toward more flexible, adaptive, and robust robot autonomy. Unlike conventional software, which executes a rigid set of programmed instructions, AI agents  are designed to 
understand complex information, reason about it, and generate a sequence of actions to perform specific tasks \cite{Brown2020, Vaswani2017}. 
This approach fundamentally shifts the development paradigm from imperative programming, where every step of a task is explicitly coded (e.g., \textit{move forward 1 meter, turn left 90 degrees}), to declarative goal setting, allowing the agent to autonomously determine how to fulfill its mission (e.g., \textit{inspect the valve}). By leveraging vast pre-trained knowledge repositories, AI agents exhibit emergent reasoning capabilities that can adapt to new tasks and conditions without requiring exhaustive manual reprogramming \cite{Laird2012, Anderson2004}. 
However, the integration of these agents in robotic applications introduces new challenges, such as ensuring robustness and verifiability of the decision made and protecting against possible \textit{hallucinations} \cite{Ji2023, Huang2023}. The increasing capabilities of advanced AI systems also bring to the forefront critical considerations related to AI safety and verification, encompassing complex issues such as alignment of the mental models of the operators and the AI system \cite{shen2023large}. 

UROSA is inspired by ROSA \cite{nasa}, which is used as a parser from high level natural language to generate actions through ROS messages. Unlike ROSA, our work integrates agentic AI at multiple levels of the autonomy architecture, able to make decisions as well as communicate through ROS. This is a novel and highly capable autonomy framework that fundamentally \emph{replaces the traditional paradigm of a human-governed main program with a distributed network of specialised and dynamically adaptable AI agents}. This distributed architecture represents a radical shift towards true autonomy, where once a mission is initiated with high-level commands, the system operates with minimal human interaction and oversight. 
The distributed AI agents communicate and solve problems together, without needing human intervention or the development of coding or extra programmatic loops. 
This allows for emergent system-level intelligence, where human interaction primarily involves providing high-level commands through natural language. Each agent is responsible for a specific aspect of the robot’s operational workflow, ranging from multimodal perception (e.g., vision, depth, sonar) to high-level strategic planning. Crucially, the AI agents within UROSA are not merely generative; they are designed as \textbf{agentic AI entities}, capable of autonomous decision-making and action-taking within their environment without requiring continuous human intervention in the loop \cite{durante2024agent}.

The realisation of this distributed cognitive architecture is founded upon the following key innovations that form the core of our UROSA framework:

\begin{itemize}
    \item \textbf{Decoupled Reasoning and Environmental Adaptability}: The framework achieves architectural flexibility by replacing traditional code-based logic with pretrained AI agents that handle system functions. Domain experts, for example a Remotely Operated Vehicle (ROV) pilot, can build systems using these existing agents without extensive reengineering. The system is adaptable in near-real time to a large variety of changes in the environment and internal states, and can access a much larger set of data through its ability to understand descriptions in natural language accessible from a variety of sources, e.g., live meteorological ocean data, vehicle design. %
    \item \textbf{Behaviour adaptation and lifelong learning}: The system is also able to learn and adapt in real-time to improve existing agentic behaviour or generate new ones, based on prompts from operators or other agents.  It can also improve over time by using previous experiences and operator feedback. This flexibility is supported by a Vector Database (VDB) that stores and retrieves past experiences, observational data, simulator outcomes and external knowledge. This comprehensive data forms the foundation for Retrieval-Augmented Generation (RAG), facilitating context-driven and informed decision-making.
    \item \textbf{Autonomous On-the-Fly Function Extension}: The system can generate code on the fly to deal with unforeseen circumstances to extend its functionality dynamically based on real-time requirements identified autonomously by the Planning Agent. This means that if a new functional component is needed for the mission, the Planning Agent can request its generation, testing, and integration without human code intervention. These capabilities can be further refined through online, agent-driven instructional tuning, where agents learn optimal response strategies through iterative, interactive feedback.
    The new code can be validated in simulation before being deployed on the real platform. This enables unprecedented adaptability in response to evolving mission demands and environmental contingencies. 
    \item \textbf{Dynamic, Predictive System Diagnostics}: The system is able to perform diagnostics of its internal state and environmental conditions, without the requirement of a predefined static fault tree or a fixed set of unit tests. 
    \item \textbf{Inherent Safety and Control}: UROSA explicitly addresses aspects of AI safety and control. Through the introduction of several robust ROS~2 mechanisms and architectural constraints, such as a dedicated safety parser and a layered design, we introduce fine-grained control over the agents' outputs and behaviours. This framework aims to mitigate hallucinations and increase the likelihood of actions aligning with predefined safety protocols, thereby contributing to the critical goal of aligning AI with human values in autonomous operations. 
\end{itemize}

While UROSA leverages contemporary LLMs and VLMs, its foundational architecture is inherently model-agnostic and designed with a forward-looking perspective. Recognising the rapid advancements in artificial intelligence 
its modular design allows for the seamless integration of future, more advanced AI paradigms as they emerge. 
We aim to propose a general concept for distributed cognitive autonomy that can evolve and benefit from future breakthroughs in AI, ensuring its continued relevance and adaptability.

We validate each of the key innovations previously outlined through a series of demanding use-case scenarios. These empirical tests provide concrete evidence of the system's performance in areas such as real-time adaptation, autonomous code generation, continuous learning, and advanced diagnostics. To further illustrate these capabilities, this paper is accompanied by a supplementary video\footnote{The video is available at: \url{https://markusbuchholz.github.io/urosa.html}} demonstrating our key experiments in both simulation and real-world deployments. 
Finally, we analyse these results to address the critical challenges of AI safety and verifiability, and discuss the future trajectory of distributed cognitive architectures like UROSA in an era of rapidly advancing artificial intelligence.

\section{Literature Overview: The Ascent of Cognitive Autonomy in Robotics}
\label{sec:literature}

The pursuit of cognitive autonomy has historically navigated a path between symbolic reasoning and reactive, embodied intelligence. Early endeavours, influenced by symbolic AI, sought to replicate cognition through formal logic and planning \cite{McCarthy1959, Nilsson1984, FikesNilsson1971}, but these top-down systems often proved unstable when faced with the uncertainty and dynamism of the real world \cite{Brooks1991, McCarthyHayes1969, AgreChapman1987}. In response, a bottom-up approach emerged with behaviour-based robotics and the principle of embodied, situated cognition, which emphasised that intelligence arises from the direct sensorimotor interaction between an agent and its environment \cite{Brooks1986, PfeiferScheier1999, Clark1997, Wilson2002}. While architectures like subsumption produced remarkably robust reactive behaviours \cite{Arkin1998}, they lacked the capacity for abstract reasoning and complex planning. Cognitive architectures such as SOAR \cite{Laird2012} and ACT-R \cite{Anderson2004, Anderson2007} attempted to bridge this gap by integrating symbolic processing with more plausible cognitive mechanisms, yet often struggled with the knowledge acquisition bottleneck and extensive hand-engineering required for novel domains \cite{BrachmanLevesque1984}.

The recent advent of LLMs has been a paradigm shift, offering powerful new avenues for high-level robotic cognition \cite{Kumar2024-tz}. Trained on vast datasets, models like GPT-4 have demonstrated remarkable emergent capabilities in reasoning and natural language understanding \cite{Brown2020, Vaswani2017, OpenAI2023, Bubeck2023}, making them potent tools for robotics \cite{Mirchev2023}. Research has rapidly demonstrated their efficacy in translating high-level natural language commands into actionable plans, as seen in works like SayCan \cite{Ahn2022, Brohan2023}. This has been extended by grounding language in rich perceptual data using VLMs \cite{Radford2021, Gupta2023, Driess2023, Jia2021, Zhu2023}, generating executable robot code from text \cite{Liang2023, Zeng2023}, and leveraging interactive reasoning paradigms such as ReAct \cite{Yao2023} for complex, embodied problem-solving.

However, integrating LLMs effectively and safely into robotic systems introduces a new set of research challenges, moving the frontier from single-model planning to robust, multi-agent architectures. The concept of a single, monolithic \textit{robot brain} gives way to distributed cognitive architectures, which promise greater modularity, fault tolerance, and emergent intelligence from the coordination of multiple specialised agents \cite{Cao1997, Stone2000, Wooldridge2009, OlfatiSaber2007}. Realising this vision of true cognitive autonomy \cite{Bradshaw2017, Longo2023, BeerRandallFitch2014} requires overcoming significant hurdles. These include ensuring robustness against model \textit{hallucinations} \cite{Ji2023, Huang2023, Maynez2020}, achieving real-time performance on resource-constrained hardware \cite{VerhelstMoons2017}, and addressing the critical challenges of AI safety, verification, and ethical alignment in safety-critical applications \cite{Amodei2016, KoopmanWagner2017, Lin2011, WallachAllen2008}.

This paper introduces UROSA, a novel distributed AI agent architecture designed to directly address this new frontier. Our work moves beyond using a single LLM for planning and instead proposes a collaborative ecosystem of specialised agentic nodes integrated deeply within the ROS~2 framework. By decentralising cognition, UROSA enhances robustness and scalability. It explicitly tackles the challenges of reliability and safety through mechanisms such as RAG for factual grounding, per-agent safety parsers to validate outputs, and dynamic, on-the-fly function generation to adapt to unforeseen circumstances. Through the design and extensive empirical validation of this framework, we demonstrate a concrete and scalable pathway toward more capable, adaptable, and reliable cognitive autonomy in real-world robotic systems.

\section{Architecture}
\label{sec:architecture}

\begin{figure*}[tp]
    \centering
      \includegraphics[width=0.95\textwidth]{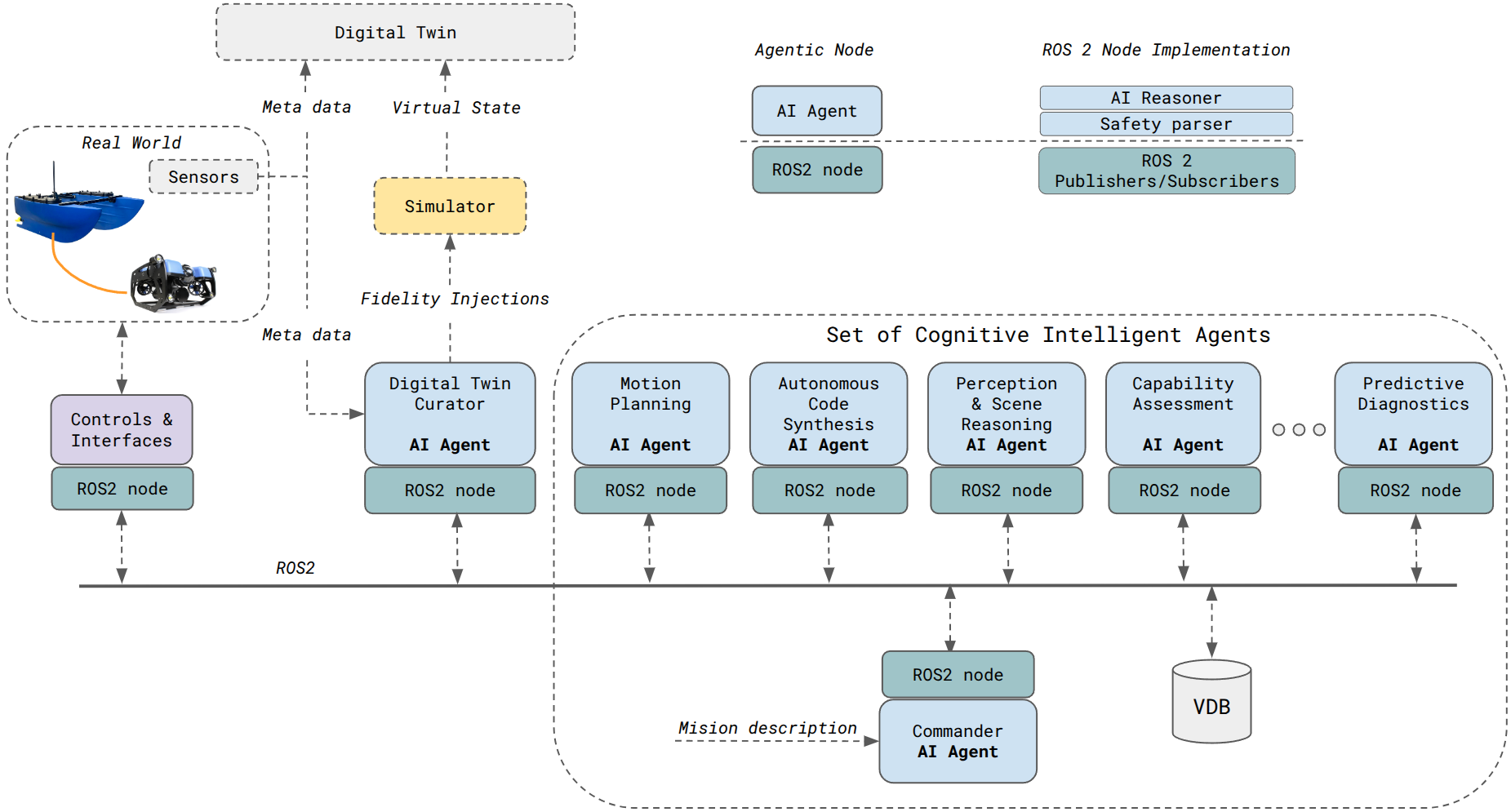}
    \caption{The UROSA Cognitive Architecture. The system is comprised of three main parts: (left) the real-world vehicle and its low-level controls; (top) the Digital Twin environment, which is continuously updated by a dedicated curator agent; and (center) the cognitive core of the framework, a distributed set of intelligent AI agents. The \textbf{Commander AI Agent} receives the high-level mission description and acts as a central orchestrator, ensuring cognitive consistency. It directs a crew of \textbf{Specialist Agents}, each responsible for a specific domain like planning, perception, or diagnostics. All agents communicate over the ROS~2 bus and access a shared, distributed \textbf{VDB} for long-term memory and experiential learning. The \textbf{Digital Twin Curator Agent} observes real-world sensor data \textit{Meta data} and injects updates \textit{Fidelity Injections} into the \textit{Simulator}, whose \textit{Virtual State} is fused with real-world data to form the comprehensive Digital Twin. The inset (top right) details the composition of each \textbf{Agentic Node}, which fuses the high-level \textbf{AI Agent} with its \textbf{ROS 2 Node Implementation}, encapsulating the \textbf{AI Reasoner}, a \textbf{Safety Parser}, and communication interfaces.}
    \label{fig:arch}
\end{figure*}

The UROSA framework fundamentally reimagines robotic autonomy by replacing a single, monolithic control program with a distributed cognitive architecture, as depicted in the overview in Fig.~\ref{fig:arch}. This design is rooted in the principles of embodied and situated cognition~\cite{PfeiferScheier1999, Clark1997, Wilson2002}, where intelligence emerges from the dynamic interaction between multiple specialised agents and their environment. At its core, UROSA decentralizes high-level reasoning across a network of AI agents, each functioning as an intelligent, collaborative unit. These agents leverage the robust communication backbone of ROS~2 to coordinate their actions and share knowledge. This hybrid architecture follows a clear separation: deterministic, low-level controllers handle time-critical tasks like stabilization, while the distributed AI agents perform the high-level cognitive functions of reasoning, planning, and adapting to novel situations.

The fundamental building block of this architecture is the \textit{Agentic ROS~2 Node}. As detailed at the top right of Fig.~\ref{fig:arch}, this is not a standard ROS~2 node but a composite entity. It fuses the high-level reasoning of an \textit{AI Agent} with the robust communication of its ROS~2 Node Implementation. Each of these nodes encapsulates the core \textbf{AI Reasoner}, a critical \textbf{Safety Parser} to validate all outputs, and standard \textbf{Publishers/Subscribers} for system-wide communication. This modular design is a core innovation of UROSA, allowing specialised intelligence to be embedded directly within the robotics framework.

\subsection{The Set of Cognitive Intelligent Agents}
The cognitive core of UROSA is not a single \textit{brain} but a distributed team of agents with distinct roles, all communicating over ROS~2.

\paragraph{The Commander AI Agent}
This agent serves as the central cognitive orchestrator, akin to a vessel's captain. Upon receiving a high-level mission description, it utilises techniques inspired by \textit{Chain-of-Thought} prompting to decompose the goal into a coherent sequence of sub-tasks for the specialist agents~\cite{Wei2022CoT}. Its primary role is to maintain cognitive consistency by employing a reflection and revision cycle, similar to the \textit{self-correction} processes described in state-of-the-art agent architectures~\cite{Madaan2023SelfRefine}.

\paragraph{The Autonomous Crew (Specialist Agents)}
A suite of specialist agents, analogous to an expert crew, handles specific functional domains. The agents shown in Fig.~\ref{fig:arch} are key examples, including the \textbf{Perception \& Scene Reasoning Agent}, the \textbf{Motion Planning Agent}, the \textbf{Autonomous Code Synthesis Agent}, the \textbf{Predictive Diagnostics Agent}, the \textbf{Capability Assessment Agent}, and the \textbf{Digital Twin Curator Agent}. This set is not fixed; the modularity of the framework allows for the seamless integration of new agents as mission complexity demands.

\paragraph{Shared Resources (ROS~2 \& VDB)}
All agents are connected by the ROS~2 framework, which serves as the distributed \textbf{communication backbone} for the entire system. Furthermore, the VDB serves as a distributed, long-term memory accessible by all agents, enabling them to learn and improve their performance over time through RAG.

\subsection{Operational Flow: Digital Twin and Real-World Interaction}
A typical mission unfolds in two interconnected loops:

\textbf{1. Cognitive Control Loop:} The \textit{Commander Agent} receives a mission and directs the specialist agents to generate plans and actions. For instance, the \textit{Motion Planning Agent} generates a trajectory, which is passed to the low-level \textit{Controls \& Interfaces} to be executed by the robot in the \textit{Real World}.

\textbf{2. Digital Twin Loop:} Concurrently, the robot's \textit{Sensors} provide real-time \textit{Meta data} to the \textit{Digital Twin Curator Agent}. This agent reasons about the real-world state and sends \textit{Fidelity Injections}-updates and corrections-to the \textit{Simulator}. The simulator's \textit{Virtual State}, now synchronised with reality, is used to create a high-fidelity \textit{Digital Twin} for predictive analysis and proactive planning.

\section{Mechanisms}

This section provides a detailed description of UROSA's main mechanisms, which were initially introduced in Section I. Section V will then present different use cases that demonstrate the performance of each mechanism.

\subsection{Decoupled Reasoning and Environmental Adaptability}
A primary innovation of UROSA is its ability to decouple high-level mission goals from low-level code implementation. This is achieved not just by distributing agents, but by fundamentally replacing traditional programmatic logic with a structured, pre-configured reasoning process embedded within each AI agent. 

The other innovation is the adaptability that these agents allow. The main engine behind UROSA's adaptability is the pre-deployment \textit{tuning} of each agent for its specific role. This is achieved not by changing model weights, but by engineering a detailed set of instructions within the \texttt{SYSTEM prompt} of its \textit{Modelfile}. This prompt acts as the agent's core cognitive and behavioural model, containing:
\begin{itemize}
    \item \textbf{A Core Directive:} Defining its primary function and role within the multi-agent system (e.g., \textit{You are a motion planner} or \textit{You are a follower agent that must coordinate with a master agent}).
    \item \textbf{Domain Knowledge and Physical Models:} Textual descriptions of vehicle dynamics, environmental forces, and physical constraints like tether properties.
    \item \textbf{Reasoning Guidelines:} Instructions on how to approach a problem, covering both individual and cooperative logic (e.g., \textit{You must compute a collision-free path that yields to the master agent's trajectory}).
    \item \textbf{A Strict Output Format:} A rigid template for its response, as detailed in our safety strategy (Section~\ref{sec:safety})
\end{itemize}

In operation, the agent's ROS~2 node subscribes to relevant topics and passes this multi-modal information to the tuned AI core. The AI agent then executes its embedded reasoning process to formulate a solution. This entire \textit{thinking process}, guided by the initial \texttt{SYSTEM prompt}, replaces thousands of lines of explicit code, allowing for unprecedented flexibility.

\subsection{Behaviour adaptation and lifelong learning}
UROSA is designed for continuous learning, where agents adapt their behaviour based on new information and past experiences. We 
enable this capability through two distinct, complementary mechanisms: (1)  learning using a VDB with RAG for knowledge-grounding, and (2) online policy refinement using a novel \textit{Teacher-Student} model for behavioural shaping.

The primary mechanism for grounding and learning from past data is the VDB. At the start of a mission, a \textbf{Flexible AI Agent} is instantiated with a specific behaviour defined in its \texttt{SYSTEM prompt}. This initial prompt instructs the agent on its core task and, crucially, how to utilise information from the VDB. During operation, the VDB is dynamically updated with valuable, curated data. In each operational cycle, the agent queries the VDB to retrieve contextually relevant past experiences. This RAG process grounds the agent's reasoning, dramatically improving decision quality and accelerating performance by leveraging a growing repository of relevant experience.

To facilitate more direct online behavioural refinement, we introduce a novel \textbf{Teacher-Student Instructional Tuning} mechanism. This approach enables autonomous adaptation of a Student agent's core policy through direct interaction with a Teacher agent. The process unfolds over episodes: the Student performs an action (e.g., generates a description). The Teacher agent analyses the response and its corrective action is to generate a new, modified \texttt{SYSTEM prompt} for the Student. This prompt acts as a rich, instructive policy guidance signal. This new prompt is then used to create a new, re-tuned instance of the Student agent for the next episode, creating a powerful meta-learning loop where the Teacher actively shapes the Student's core reasoning policy from one episode to the next.

\subsection{Autonomous On-the-Fly Function Extensibility}

A key innovation of UROSA is its ability to autonomously generate, test, and integrate new software components at runtime, extending its own functionality without direct human intervention. This capacity for self-extension allows the system to adapt its software architecture to unforeseen challenges or mission requirements, effectively enabling runtime \textit{self-repair}.

This mechanism is enabled by the \textbf{Autonomous Code Synthesis AI Agent}, operating under the direction of the \textit{Commander AI Agent}. The process begins when the \textit{Commander} autonomously identifies a functional gap in the system, for instance, determining that a specific data filter or a novel planning algorithm is required to handle the current situation. It then formulates a set of high-level requirements and sends them as a natural language request to the Code Synthesis agent.

It is crucial to note that the AI's role is strictly that of a powerful code synthesiser; it does not perform the requested task itself. Upon receiving the requirements, the agent:
\textbf{(1) Synthesises Code:} Generates the source code for a complete ROS~2 node in a suitable language like Python; \textbf{(2) Generates Tests:} Simultaneously creates a suite of unit tests to validate the new code's functional correctness and safety; \textbf{(3) Saves and Deploys:} If the automated tests pass, the generated source code is saved as a new executable file. This new node is then seamlessly launched and integrated into the live ROS~2 computation graph.
This entire workflow (from the identification of need to the deployment of a validated new capability) is performed on-the-fly.

\subsection{Dynamic, Predictive System Diagnostics}\label{sec:diagnostics_mechanism}
UROSA features an advanced diagnostics capability that moves beyond static fault trees, using an AI agent to reason about the system's health based on live data. This allows for the detection of complex, emergent failures that are not defined by simple error codes.

This capability is enabled by a dedicated \textbf{Diagnostic AI Agent} 
whose behaviour is defined by a highly structured, task-specific reasoning process embedded within its core \texttt{SYSTEM prompt}. This transforms the generic LLM into a specialised diagnostic expert. The prompt instructs the agent to:
 \textbf{(1)} Perform a multi-step time-series analysis over a sliding window of 10 consecutive JSON status messages from the vehicle.
\textbf{(2)} Utilise an embedded domain-specific physical model-the vehicle's thruster allocation matrix-to determine the expected thruster behaviour for any commanded movement. \textbf{(3)} Compare the expected PWM values against the observed data for each thruster in the time window, using defined logic to classify distinct fault types (e.g., \textit{dead} or \textit{out-of-range}). \textbf{(4)} Aggregate these findings to identify drifts, asymmetries, or other anomalies.
\textbf{(5)} Report its findings in a rigid, three-line format: a) Issue Identification, c) Status Summary, and b) Suggested Action.

This entire diagnostic procedure, governed by the initial prompt, allows the agent to use its holistic understanding of the vehicle's expected behaviour to diagnose issues without relying on a predefined list of faults.

\subsection{Inherent Safety and Control Mechanisms }
Finally, UROSA explicitly addresses AI safety and control through its architecture, aiming to ensure that the agents' actions are verifiable and aligned with human intent. This is achieved through a multi-layered safety strategy, implemented from the agent's creation to its final output.

Safety in UROSA is not a single component but an integrated, three-tiered approach:

\textbf{1. Proactive Behavioural Scaffolding:} The first and most critical layer of control is established during each agent's creation via its \textit{Modelfile}. This involves embedding a complete, task-specific reasoning process and strict output format directly into the agent's core \texttt{SYSTEM prompt}. By providing detailed instructions, domain knowledge, physical models, and reasoning guidelines, we transform the generic LLM into a highly specialised and predictable, task-oriented engine. The detailed implementation of this principle for the \textit{Diagnostic Agent}, as described in Section~\ref{sec:diagnostics_mechanism}, is a prime example of this technique. This same scaffolding approach is applied to all agents in the UROSA framework to ensure reliable and verifiable behaviour.

\textbf{2. Dynamic Contextual Grounding:} The second layer is provided by the heavy reliance on RAG, orchestrated by the \textit{Brain Agent}. While the \texttt{SYSTEM prompt} defines \textit{how} an agent should reason, RAG ensures that \textit{what} it reasons about is grounded in a verified knowledge base and real-world data from the VDB. This provides the necessary context to make its structured output factually relevant to the current mission, further mitigating the risk of ungrounded decisions.

\textbf{3. Reactive Output Validation:} As a final, reactive backstop, every agentic node contains a \textbf{Safety Parser}. This component validates all LLM outputs before they are published as ROS~2 messages. It acts as a critical final checkpoint, checking for correct syntax and ensuring the command adheres to predefined operational safety rules (e.g., maximum speed or depth). Any command that fails this validation is blocked.

Furthermore, the distributed architecture itself provides inherent fault tolerance, as the failure of one specialist agent does not necessarily compromise the entire system.

\section{Results}
To substantiate the claims made in this paper, we conducted a comprehensive evaluation strategy designed to validate each of UROSA's core innovations. The experiments were both in simulation, using in a high-fidelity simulation environment, and in real-time robotic platforms. We utilise an Autonomous Surface Vehicle (ASV) and an Autonomous Underwater Vehicle (AUV)  \cite{willners2021market}, in a water tank (3.5\,m x 3.0\,m x 2.5\,m) available in our laboratories.  The following subsections are structured to address each innovation introduced in Section~I and described in detail in Section~IV, presenting empirical validation through targeted use-case scenarios.

\subsection{Decoupled Reasoning and Environmental Adaptability}

We validated this through several complex scenarios, covering both multi-agent coordination and single-robot manipulation, with quantitative results presented for each to demonstrate performance and reliability.

\subsubsection{Constrained Multi-Robot Coordination With Obstacles}
\begin{figure}[t]
\centering
\includegraphics[width=0.5\textwidth]{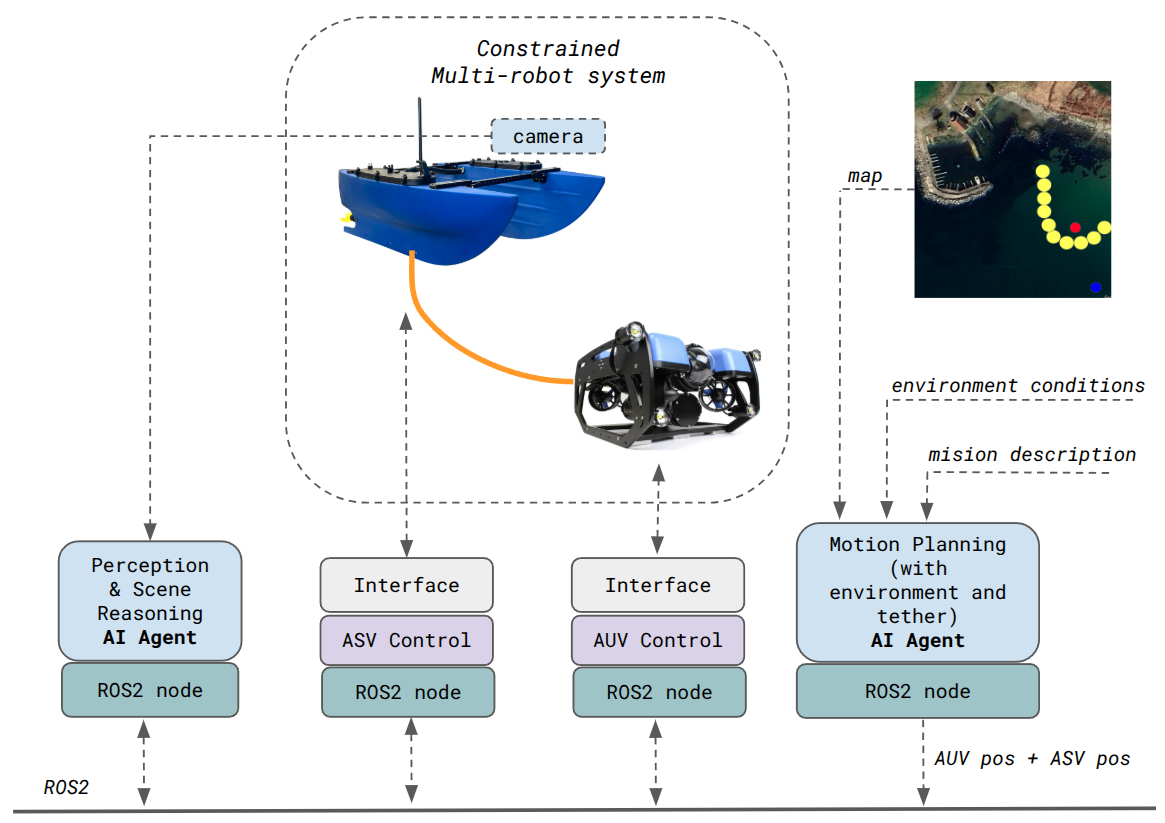}
\caption{Illustration of UROSA's flexible, multi-modal planning architecture. The system's AI agents act as a cognitive hub, integrating pre-defined mission data or dynamic, real-time perceptions to enable seamless switching between goal-driven and perception-driven behaviors. In the map-based planning scenario shown, the agent must generate a collision-free path from the \textbf{blue circle} (start position) to the \textbf{red circle} (goal position), treating the \textbf{yellow circles} as obstacles.}
\label{fig:constrained_multi_robot_dynamic}
\end{figure}

In this simulation, a tethered ASV-AUV system (with a 10\,m tether) was tasked with achieving a position goal while avoiding an obstacle, as shown in Fig.~\ref{fig:constrained_multi_robot_dynamic}. This required the AI agent to reason about the complex system dynamics, including the tether's catenary shape, to maintain a valid formation throughout the maneuver. To benchmark UROSA's performance, we compared it against a traditional motion planner based on an enhanced A* algorithm~\cite{buchholz2025framework, buchholz2025tethered}. A critical distinction in this evaluation is the source of obstacle information: the traditional A* planner was provided with the ground-truth position of the obstacle, whereas the UROSA framework had to rely on a \textit{Perception \& Scene Reasoning Agent} to detect and track the obstacle from a live camera feed.

The validation was performed across 5 distinct mission configurations, each featuring a unique map with different start positions, goal locations, and obstacle layouts to test the system's adaptability. For each of these configurations, we conducted 5 trials to account for any stochasticity in the perception and planning process. The key comparative metrics-the additional positioning error introduced by the AI agent compared to the A* baseline, and the agent's task success rate-are presented in Table~\ref{tab:constrained_coord_comparison}. The results show an expected trade-off: the AI agent's positioning error is higher because it operates on a live perception feed, unlike the A* planner, which used perfect information. Nevertheless, achieving a success rate of up to 80\% on this difficult perception-in-the-loop task is a significant result, demonstrating that the framework can close the loop from raw perception to complex multi-robot coordination using only emergent reasoning.

\begin{table}[t]
\centering
\caption{Dynamic Avoidance: AI Agent vs. Enhanced A*}
\label{tab:constrained_coord_comparison}
\begin{tabular}{c|cc}
\hline
\textbf{Mission ID} & \textbf{Error Delta (AI - A*) (m)} & \textbf{AI Agent Success (\%)} \\ \hline
1 & 2.56 & 80\% \\
2 & 3.78 & 80\% \\
3 & 4.23 & 80\% \\
4 & 6.37 & 60\% \\
5 & 4.68 & 60\% \\ \hline
\end{tabular}
\end{table}

\subsubsection{Map-Based Path Planning for Multi-Robot Systems} 
To test the ability to interpret abstract data, we tasked the AI system with planning a path for the tethered ASV-AUV system using only a 2D map image and a text-based goal (Fig.~\ref{fig:constrained_multi_robot_dynamic}). We again compared the AI-driven approach to the enhanced A* planner~\cite{buchholz2025framework, buchholz2025tethered}. The averaged results over 5 trials for each map layout are compared in Table~\ref{tab:map_planning_comparison}.

\begin{table}[t]
\centering
\caption{Map-Based Planning: AI Agent vs. Enhanced A*}
\label{tab:map_planning_comparison}
\begin{tabular}{c|cc}
\hline
\textbf{Map ID} & \textbf{Error Delta (AI - A*) (m)} & \textbf{AI Agent Success (\%)} \\ \hline
1 & 3.34 & 80\% \\
2 & 2.98 & 80\% \\
3 & 7.81 & 60\% \\
4 & 8.13 & 60\% \\
5 & 6.26 & 40\% \\ \hline
\end{tabular}
\end{table}

The results in Table~\ref{tab:map_planning_comparison} demonstrate the trade-offs of an end-to-end reasoning system. The higher positioning error and variable success rates are an expected consequence of the task's complexity; the AI agent must first interpret the raw map image, identify obstacles, and then plan a path, a process with more inherent uncertainty than the A* planner's, which operates on pre-defined maps. The key achievement is that the system can succeed in this complex task at all, reaching success rates of up to 80\% on. This is a significant leap in autonomy because the UROSA framework reasons directly from raw images and text to create a plan, bypassing the time-consuming manual setup required by traditional planners.

\subsubsection{Flexible Motion Planning for a UVMS}
\begin{figure*}[t]
\centering
\includegraphics[height=4cm]{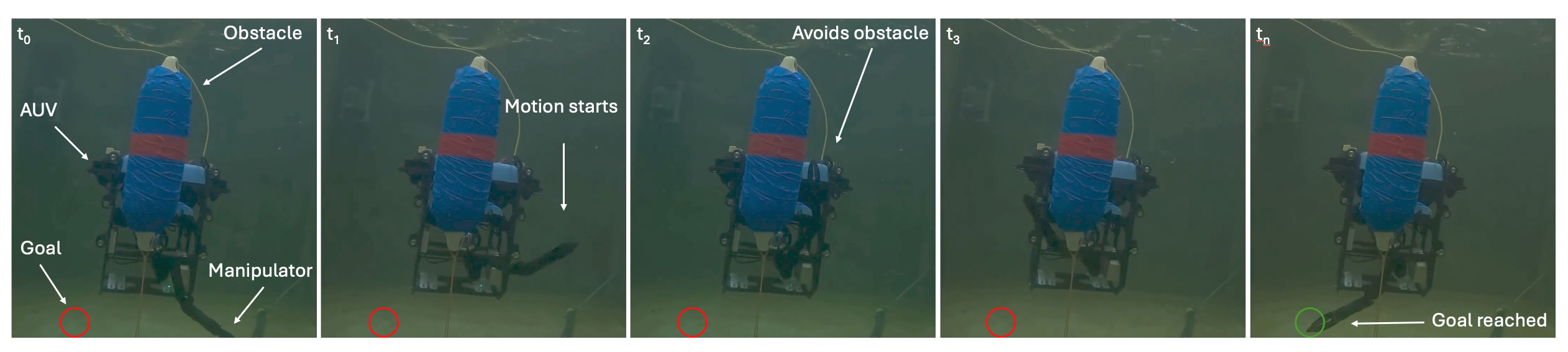}
\caption{The AI agent workflow for the UVMS manipulation task. A high-level textual command from the \textit{User terminal} is interpreted by the \textit{Commander AI Agent}, which then directs the \textit{Motion Planning} and \textit{Perception} agents to execute the task.}
\label{fig:uvms_workflow}
\end{figure*}

We further validated this principle on an Underwater Vehicle Manipulator System (UVMS) \cite{buchholz2025contextaware,morgan2022autonomous}, focusing on autonomous motion planning driven by textual commands (Fig.~\ref{fig:uvms_workflow}). The experiment, conducted in a real-world tank, was designed to test the \textit{Commander AI Agent}'s ability to interpret nuanced natural language instructions. To simplify the test's scope and focus purely on command interpretation and planning, the coordinates for key locations like \textit{goal 1} and \textit{goal 2} were predefined within the agent's \texttt{SYSTEM prompt}.

To provide a quantitative benchmark, we compared its performance against a baseline traditional planner using a naive A* search on a Probabilistic Roadmap (PRM). To test the reliability of the AI's language understanding, we conducted 10 trials using prompts with varied phrasing and intent. These included explicit instructions for avoidance (e.g., \textit{Go to goal 1 and check the obstacles on the way}) as well as more ambiguous phrasing (e.g., \textit{Inspect goal 2 and check for any boxes on your way}). Crucially, the test also included direct commands that omitted any mention of obstacle avoidance, such as \textit{Inspect goal 2}. In these cases, to test literal command adherence, the \textit{Commander AI Agent} intentionally disregarded perceptual data about obstacles and commanded a direct path, leading to a planned collision. The averaged results of these trials are summarized in Table~\ref{tab:uvms_planning_comparison}.

\begin{table}[t]
\centering
\caption{Manipulator Planning \& Interpretation Performance}
\label{tab:uvms_planning_comparison}
\begin{tabular}{l|cc}
\hline
\textbf{Metric} & \textbf{AI Agent} & \textbf{Naive A*+PRM} \\ \hline
Planning Time (s) & 1.3 & 0.05 \\
Interpretation Success (\%) & 90\% & N/A \\
Planning Success (\%) & 100\% & 100\% \\ \hline
\end{tabular}
\end{table}

The results highlight the unique capabilities of the UROSA approach. The system achieved a 90\% \textit{Interpretation Success} rate, correctly understanding the user's intent in 9 out of the 10 trials. Crucially, for every prompt that was correctly understood, the subsequent \textit{Planning Success} rate for generating a valid trajectory was 100\%. This clearly distinguishes between the agent's highly robust (but not infallible) natural language understanding and its near-perfect planning capability once the goal is known—a distinction the traditional method (N/A) cannot address. This showcases the framework's power in translating high-level, human-like instructions directly into complex robot actions.

\begin{figure*}[t]
\centering
\includegraphics[width = \textwidth]{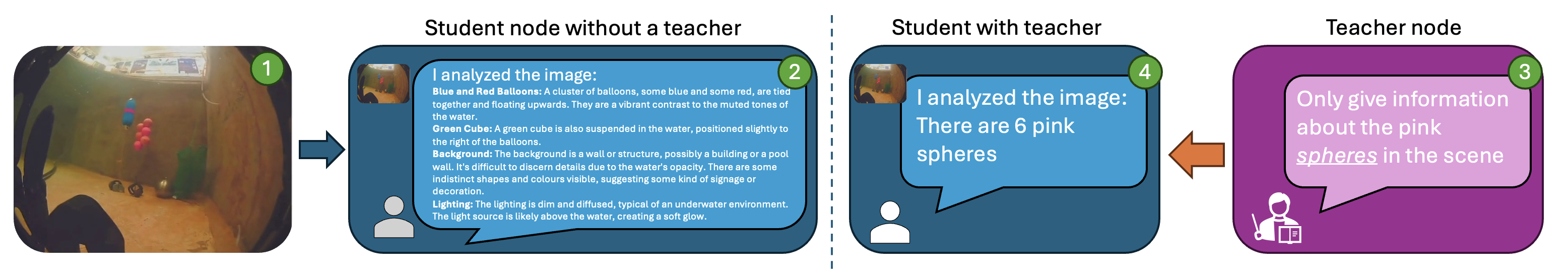}
\caption{A direct \textit{Before and After} comparison of the Teacher-Student instructional tuning mechanism. 
        (Left) Without the \textit{Teacher} enabled, the \textit{Student} agent observers the scene (1) and defaults to a verbose scene description as seen in (2). 
        (Right) With the \textit{Teacher} enabled, the Teacher gives precise feedback (3) to the \textit{Student} over several episodes, resulting in a converged output that is concise (4) and provides only the information requested by the \textit{Teacher}}
\label{fig:uvms_teach_student}
\end{figure*}

\subsection{Behaviour Adaptation and Lifelong Learning}

We evaluated these adaptive learning capabilities through targeted experiments demonstrating both experiential learning via RAG and online policy refinement.

\subsubsection{Visual Positioning with Experiential Learning}
\begin{figure}[t]
\centering
\includegraphics[width=0.47\textwidth]{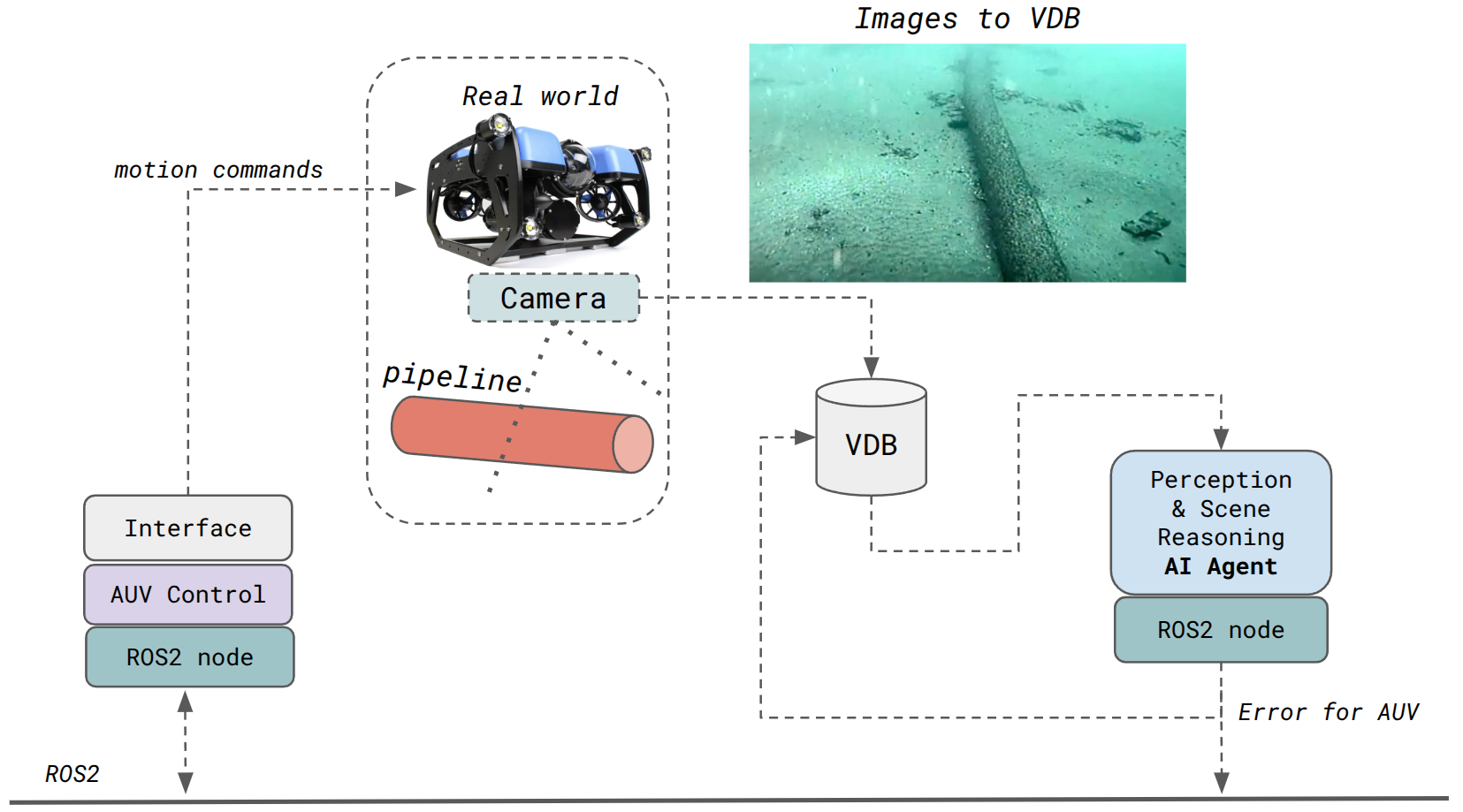}
\caption{The control loop for visually-guided disturbance rejection. The AUV's camera feeds a continuous stream of images into a VDB, which provides visual-temporal memory. The \textit{Perception \& Scene Reasoning AI Agent} uses this historical context to calculate a precise positional error, which the AUV's control system then uses to issue corrective motion commands.} 
\label{fig:pipe_tracking}
\end{figure}

To validate how the framework's use of experience improves resilience against external disturbances, we conducted simulation tests focused on visual positioning. In this scenario, an AUV was tasked with maintaining a stable position and orientation relative to a static feature on the seabed (a pipeline segment), as shown in the simulated views in Fig.~\ref{fig:pipe_tracking}. This capability is enabled by a \textbf{Flexible AI Agent} that uses a VDB as a short-term visual-temporal memory. This allows the agent to reason not just about its current state, but about the dynamics of recent events.

To isolate the impact of this experience buffer, we benchmarked the agent's performance in two conditions: (1) with VDB access disabled, and (2) with full VDB access. The test involved letting the AUV follow a pipeline and then applying a simulated external force to induce a specific lateral deviation. To evaluate the robustness of the agent's response, we tested three magnitudes of absolute deviation (1.0\,m, 1.5\,m, and 2.5\,m), applying the force to induce errors to both the left and the right. We then measured the \textit{Recovery Time}—the time taken for the agent to guide the AUV back to the pipe centerline. Each of these six conditions was run 3 times, and the results, averaged across both left and right directions for each deviation magnitude, are presented in Table~\ref{tab:pipe_tracking_comparison}.

\begin{table}[t]
\centering
\caption{Disturbance Recovery with Experiential Learning}
\label{tab:pipe_tracking_comparison}
\begin{tabular}{c|cc}
\hline
\textbf{Deviation} & \textbf{\begin{tabular}[c]{@{}c@{}}Recovery Time \\ (w/o VDB, s)\end{tabular}} & \textbf{\begin{tabular}[c]{@{}c@{}}Recovery Time \\ (w/ VDB, s)\end{tabular}} \\ 
\hline
1.0\,m & 5.9 & \textbf{2.6} \\
1.5\,m & 9.0 & \textbf{3.2} \\ 
2.5\,m & 12.6 & \textbf{4.4} \\
\hline
\end{tabular}
\end{table}

The results demonstrate the significant value of experiential learning. The agent with VDB access recovered much faster because it leveraged the visual history to infer the disturbance dynamics and issue a proactive correction command. In contrast, the baseline agent could only react to its instantaneous error, proving less effective. The key hardware-independent finding is that the VDB provides the temporal context required for more resilient control.

\subsubsection{Online Behavioural Tuning via Teacher-Student Interaction}
To demonstrate more direct online behavioural adaptation, we validated our novel \textbf{Teacher-Student Instructional Tuning mechanism}. In this paradigm, a \textit{Teacher} agent guides a \textit{Student} agent's policy by generating a new, more restrictive \texttt{SYSTEM prompt} as a form of instructive feedback, creating a powerful meta-learning loop. We conducted a series of real-world experiments in a cluttered tank environment, shown in Fig.~\ref{fig:uvms_teach_student}, where the \textit{Commander AI Agent} acted as the \textit{Teacher} and the \textit{Perception \& Scene Reasoning AI Agent} was the \textit{Student}.

The Teacher's objective was to guide the Student's policy from providing verbose descriptions of the whole scene to reporting \textit{only} the presence and location of a specific target. We ran a total of 20 trials, comprising 5 trials for each of four distinct target types: red balls, pink buoys, a fishing net, and other submerged obstacles. At each episode in a trial, the \textit{Student} generated a description, and the \textit{Teacher} provided corrective feedback by generating a new \texttt{SYSTEM prompt} to re-tune the \textit{Student's} behaviour for the next episode. The learning process was quantified over three episodes, and Table~\ref{tab:teacher_student_tuning} shows the averaged performance across all 20 trials.

\begin{table}[t]
\centering
\caption{Averaged Online Policy Refinement via Teacher-Student Loop}
\label{tab:teacher_student_tuning}
\begin{tabular}{c|cc}
\hline
\textbf{Episode} & \textbf{Avg. Response Length (Words)} & \textbf{Avg. Info. Relevance (\%)} \\ \hline
1 & 45 & 5\% \\
3 & 18 & 65\% \\
6 & 5 & 100\% \\ \hline
\end{tabular}
\end{table}

The results demonstrate a consistent and rapid convergence on the desired behaviour across all target types. On average, the Student's policy quickly shifted from general scene description to specific, targeted reporting, confirming the mechanism's ability to perform targeted, online behavioural shaping for a variety of perceptual goals using structured, linguistic guidance.

\subsubsection{Validation}
\begin{figure}[t]
    \centering
    \includegraphics[width=0.48\textwidth]{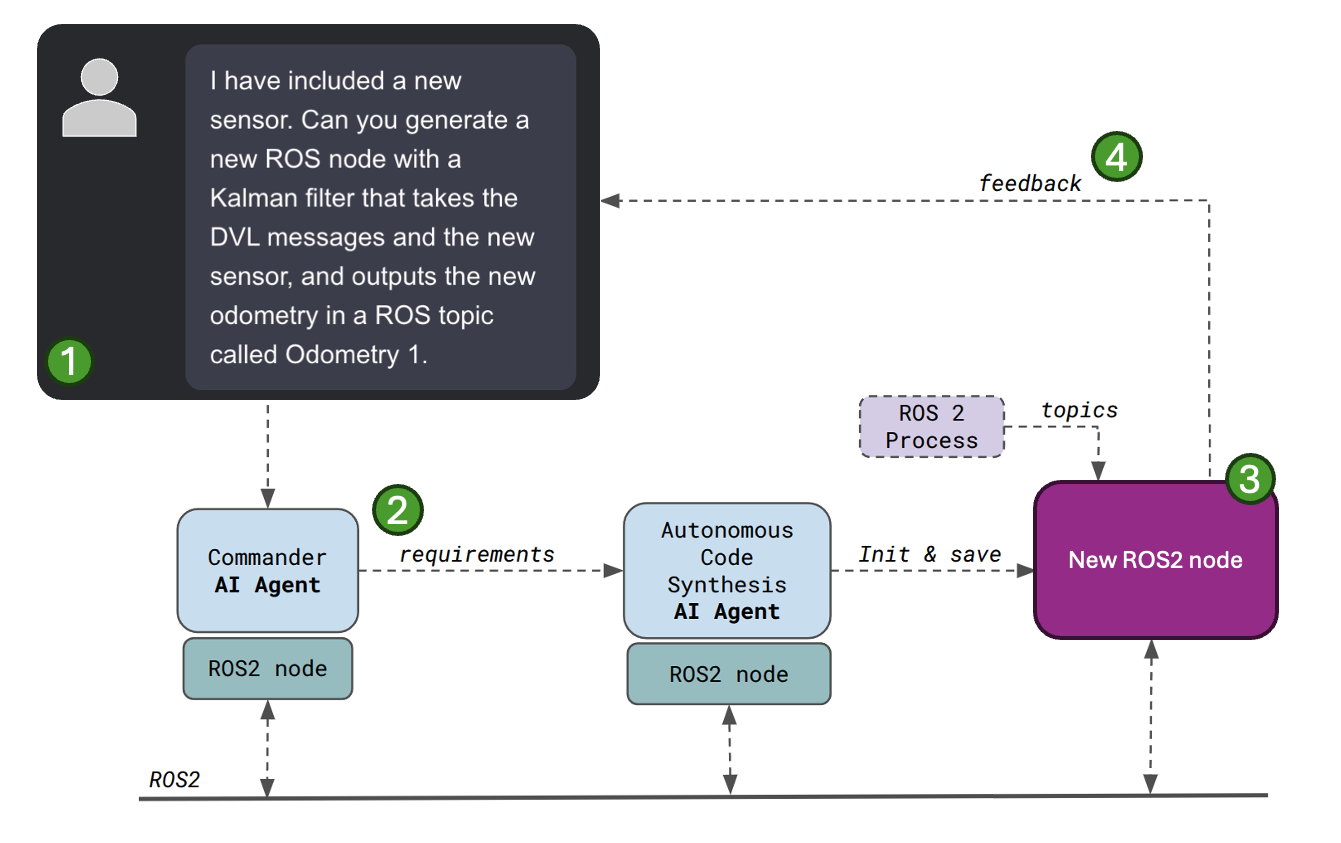}
    \caption{The evaluation workflow for autonomous node generation. 1) A user gives a query. 2) The \textit{Commander AI Agent} generates the requirements and passes them over to the \textit{Autonomous Code Synthesis Agent}. 3) The \textit{Autonomous Code Synthesis Agent} generates, tests, and deploys the new ROS 2 node. 4) For this evaluation, feedback on the new node's performance was routed to the user terminal.  It is important to note that in a fully autonomous mission, the \textit{Commander AI Agent} can make the starting request and the feedback would be sent back to it to inform subsequent decisions.}
    \label{fig:node_gen}
\end{figure}

We evaluated this mechanism's capacity for runtime software adaptation and self-repair using the controlled test setup shown in Fig.~\ref{fig:node_gen}. For this evaluation, we simulated an autonomous request from the \textit{Commander AI Agent} by manually publishing a ROS~2 topic containing the high-level natural language requirements for a new node. A trial was considered a \textit{success} only if the agent completed the full end-to-end process: correctly interpreting the request, generating valid Python code, creating and passing its own unit tests, and integrating the new node into the live system. A \textit{failure} was defined as any case where the generated code did not compile, failed its unit tests, or was functionally incorrect.

We conducted 10 trials for each of the following field-relevant scenarios:
The field-relevant scenarios included generating: \textbf{(1)} a stateful noise-reduction filter to average sensor values over a 10-sample window; \textbf{(2)} a standard sensor fusion node implementing a Kalman filter for two odometry topics; and \textbf{(3)} an emergency navigation repair node. This primary evaluation scenario simulated a critical Inertial Navigation System (INS) failure, requiring the system to synthesize a new Kalman filter to fuse compass and Doppler Velocity Log (DVL)  data to mitigate navigational drift.

The performance of the autonomous generation process is summarized in Table~\ref{tab:node_gen_results}. Most significantly, in the navigation failure scenario, the dynamically generated Kalman filter node, once integrated, successfully fused the compass and DVL data, reducing the navigational drift rate by an estimated 70\% compared to dead-reckoning with DVL alone. This provides powerful evidence of the system's ability to perform runtime self-repair of a critical capability.

\begin{figure}[t]
    \centering
    \includegraphics[width=0.48\textwidth]{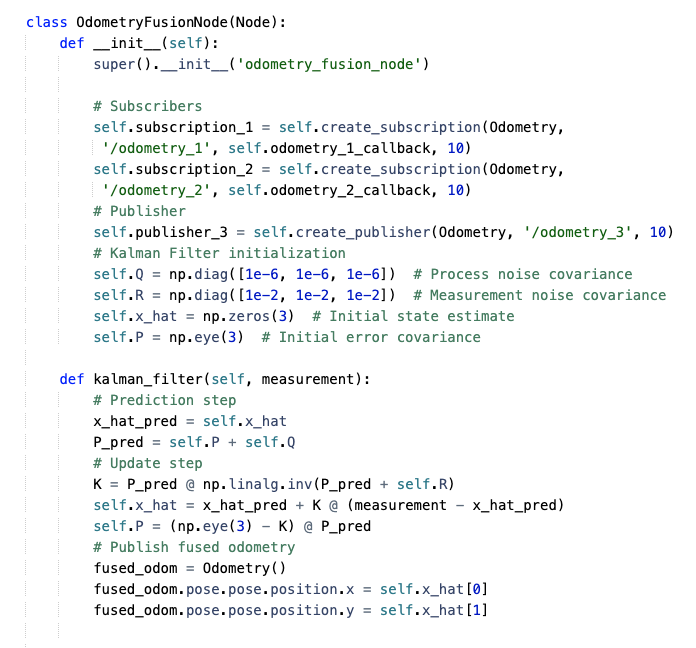}
    \caption{Excerpt of the code generated. For full details please check the website and videos.}
    \label{fig:node_gen_results}
\end{figure}

\begin{table}[t]
\centering
\caption{Performance of Autonomous Node Generation}
\label{tab:node_gen_results}
\begin{tabular}{l|cc}
\hline
\textbf{Generated Node} & \textbf{Avg. Gen. Time (s)} & \textbf{Success Rate} \\ \hline
Stateful Averaging Filter & 1.5 & 80\% \\
Kalman Filter (2x Odom) & 5.8 & 80\% \\
Kalman Filter (DVL+Compass) & 5.9 & 70\% \\ \hline
\end{tabular}
\end{table}

Most significantly, in the navigation failure scenario, the dynamically generated Kalman filter node, once integrated, successfully fused the compass and DVL data, reducing the navigational drift rate by an estimated 70\% compared to dead-reckoning with DVL alone. %

\subsection{Dynamic, Predictive System Diagnostics}
\begin{figure}[t]
    \centering
     \includegraphics[width=0.45\textwidth, height=0.27\textwidth]{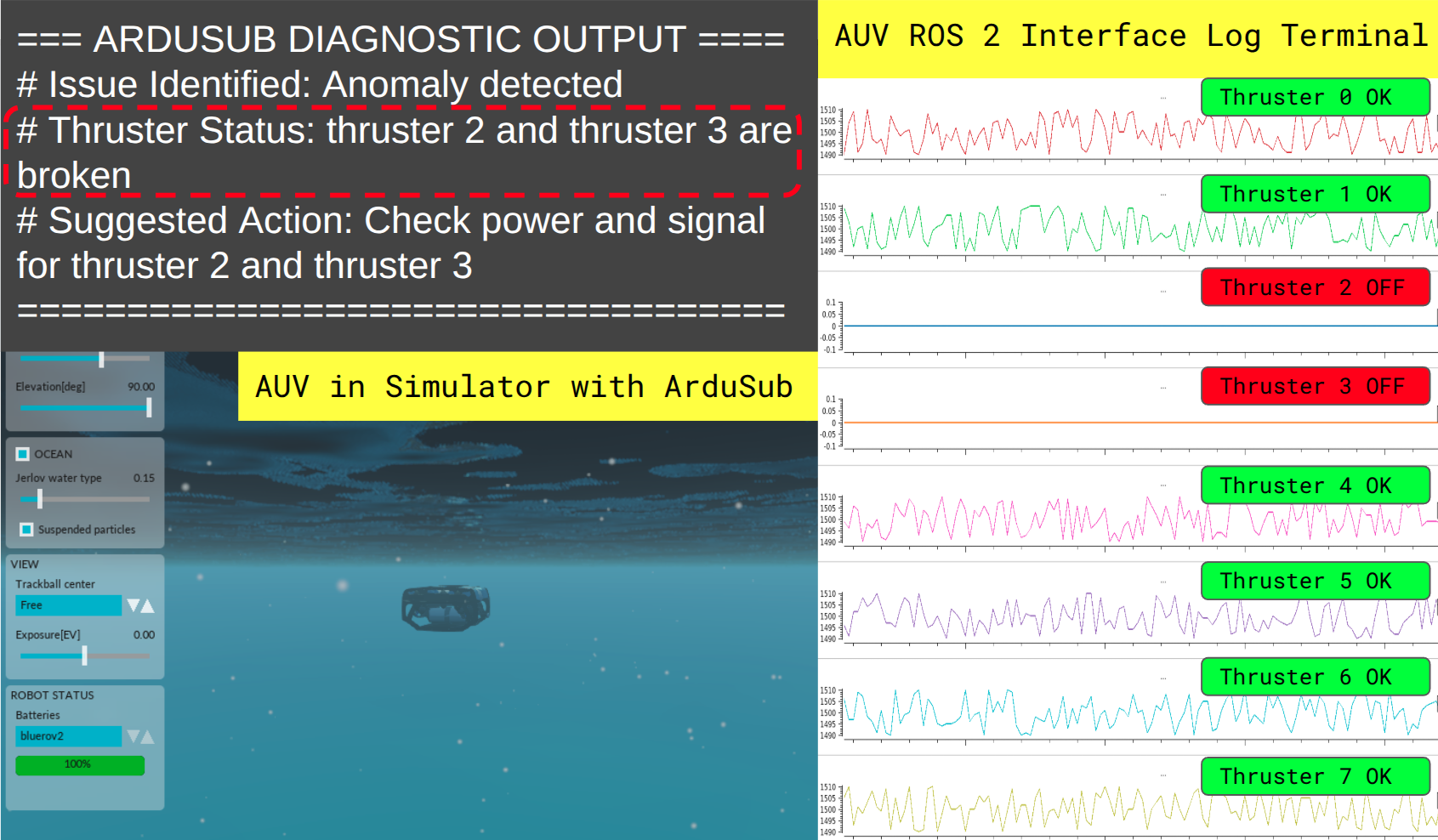} 
    \caption{Validation of the Predictive Diagnostics Agent during a simulated two-thruster failure. The plots on the right show the raw PWM signals from the ROS~2 interface, clearly indicating that thrusters 2 and 3 are unresponsive (flatlined) while the others are active. The Diagnostic Agent continuously analyzes this time-series data, reasons about the discrepancy between commanded and actual behavior, and autonomously generates the correct high-level diagnosis shown in the terminal output (top left). This demonstrates the agent's ability to translate low-level signal anomalies into a precise, human-readable fault report without relying on pre-programmed error codes.}
    \label{fig:diagnostic_transition}
\end{figure}

This capability was evaluated through a series of proof-of-concept experiments where we simulated various hardware failure modes. For each test case, we programmatically disabled specific thrusters via the vehicle's control interface. The ArduPilot system \cite{ardupilot_software} communicating with Stonefish \cite{8867434, grimaldi2025stone}, observing the lack of response, would then generate an updated vehicle status \textit{JSON} reflecting the fault. The Diagnostic AI Agent's task was to monitor this stream and correctly identify which thrusters were malfunctioning. We ran 5 trials for each fault configuration, and the agent correctly diagnosed the system state with 100\% accuracy across all tests, as summarised in Table~\ref{tab:diagnostics_results}.
To further demonstrate the agent's ability to track a dynamically changing system state, we conducted a specific transition test. First, we disabled thrusters 2 and 3 to induce a fault. After the agent correctly diagnosed this failure, we re-enabled the thrusters in the interface. The agent's subsequent report immediately and accurately reflected the system's return to a healthy state. Figure~\ref{fig:diagnostic_transition} shows the agent's real-time terminal output, capturing these precise system faults. 
\begin{table}[t]
\centering
\caption{Summary of Diagnostic Test Cases and Results}
\label{tab:diagnostics_results}
\begin{tabular}{l|l|c}
\hline
\textbf{Test Case} & \textbf{Simulated Fault} & \textbf{Diagnosis Accuracy} \\ \hline
1 & All Thrusters OK & 100\% (5/5) \\
2 & Thruster 2 Disabled & 100\% (5/5) \\
3 & Thruster 6 Disabled & 100\% (5/5) \\
4 & Thrusters 2 \& 3 Disabled & 100\% (5/5) \\
5 & Thrusters 2, 3, 6, 7 Disabled & 100\% (5/5) \\ \hline
\end{tabular}
\end{table}

\begin{figure}[t]
    \centering
    \includegraphics[width=0.45\textwidth, height=0.45\textwidth]{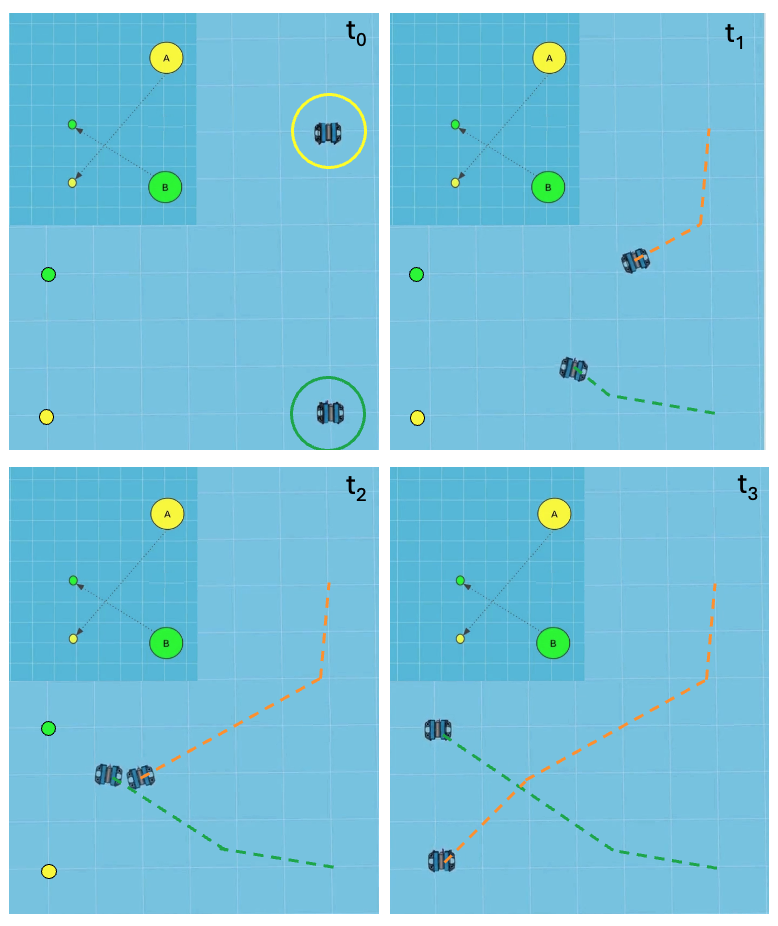} 
    \caption{Iustration of the decentralised collision avoidance scenario, where two AI agents negotiate a solution via ROS 2. Agent A, marked as Yellow, and Agent B, marked with green. Each has a starting position and a goal point, with conflicting goals. }
    \label{fig:collision_avoidance}
\end{figure}

\subsection{Inherent Safety and Control Mechanisms}
\label{sec:safety}

To validate the safety of the agent, we conduct a complex evaluation of multi-agent behaviour. Two AUVs, each controlled by an independent AI agent, navigated a maze-like environment toward a shared goal (Fig.~\ref{fig:collision_avoidance}). The agents must then coordinate to avoid each other, but must do so via emergent coordination, without preplanning. The key to this capability was embedding not only a negotiation protocol but also the vehicle's physical dimensions into each agent's behavioural constitution (\texttt{SYSTEM prompt}). This allowed each agent to reason spatially about its own footprint and the other's intended path.

The evaluation involved 8 test runs with increasingly complex trajectory conflicts, where paths crossed or required agents to pass each other in tight corridors. Each test case was run 5 times to ensure repeatability. During operation, the agents continuously share their intended trajectories via ROS~2. Each agent then predicts the future states along both paths, calculating the minimum distance between their vehicle's bounding volumes. If this predicted distance fell below a predefined safety threshold, a collision was considered imminent, triggering a human-independent negotiation dialogue. Based on its pre-configured reasoning process, one agent would identify the conflict and autonomously compute a new path yielding to the other. Table~\ref{tab:collision_avoidance} summarises the averaged performance, focusing on the efficiency of the negotiation and the safety of the resulting manoeuvre.

\begin{table}[t]
\centering
\caption{Performance of Decentralised Negotiation (Avg. over 5 runs)}
\label{tab:collision_avoidance}
\begin{tabular}{c|cc}
\hline
\textbf{Test ID} & \textbf{Negotiation Time (s)} & \textbf{Min. Safe Distance (m)} \\ \hline
1 & 0.05 & 1.2 \\
2 & 0.05 & 1.4 \\
3 & 0.06 & 2.1 \\
4 & 0.12 & 0.21 \\
5 & 0.13 & 0.22 \\
6 & 0.06 & 2.32 \\
7 & 0.14 & 0.32 \\
8 & 0.13 & 0.76 \\ \hline
\end{tabular}
\end{table}

Across all 40 trials, the agent pair successfully negotiated a collision-free path every time. The data in Table~\ref{tab:collision_avoidance} shows that the negotiation was exceptionally efficient, with a resolution consistently found in well under 0.2 seconds. It should be noted that this time is presented as a reference, as it is dependent on the specific AI model and GPU hardware used. Critically, even in the most constrained scenarios where the vehicles had to pass with very tight clearances (e.g., a minimum distance of just 0.21\,m in Test ID 4), a positive safe distance was always maintained. This consistent success in difficult configurations confirms that an effective, emergent deconfliction strategy can be derived directly from high-level textual instructions about vehicle geometry.

\section{Conclusion}
\label{sec:conclusion}
This paper presented UROSA, a distributed AI agent architecture that replaces monolithic programmatic control with a collaborative ecosystem of agentic entities. By embedding a structured reasoning process into each agent via behavioural constitutions, UROSA decouples high-level mission goals from low-level code implementation, enabling a new level of cognitive flexibility in the underwater domain. Through empirical validation, we demonstrated that this framework effectively handles multi-robot coordination from abstract inputs, improves performance through experiential learning via RAG, autonomously synthesises new software modules at runtime, and performs predictive diagnostics without static fault trees. These capabilities, protected by a multi-layered safety strategy, proved both reliable and effective across demanding scenarios.

While these findings are significant, challenges in deploying such cognitive systems remain. The engineering of effective behavioural constitutions is a complex new skill, and the formal verification and real-time performance of every AI-driven decision require further research. Future work will address these challenges by exploring two frontiers: automating the generation of agent constitutions, and enabling runtime self-reconfiguration. This latter capability would allow an agent, upon receiving a diagnosis of hardware failure, to autonomously rewrite and deploy a new control model—such as a modified thruster allocation matrix—to ensure the system continues its mission with gracefully degraded performance. This research provides a scalable and safe framework for developing more adaptable and resilient autonomous systems, moving beyond pre-programmed tools towards reasoning partners capable of overcoming both environmental and internal, unforeseen challenges.

\bibliographystyle{IEEEtran}
\bibliography{reference}

\begin{thebibliography}{10}
\providecommand{\url}[1]{#1}
\csname url@samestyle\endcsname
\providecommand{\newblock}{\relax}
\providecommand{\bibinfo}[2]{#2}
\providecommand{\BIBentrySTDinterwordspacing}{\spaceskip=0pt\relax}
\providecommand{\BIBentryALTinterwordstretchfactor}{4}
\providecommand{\BIBentryALTinterwordspacing}{\spaceskip=\fontdimen2\font plus
\BIBentryALTinterwordstretchfactor\fontdimen3\font minus \fontdimen4\font\relax}
\providecommand{\BIBforeignlanguage}[2]{{%
\expandafter\ifx\csname l@#1\endcsname\relax
\typeout{** WARNING: IEEEtran.bst: No hyphenation pattern has been}%
\typeout{** loaded for the language `#1'. Using the pattern for}%
\typeout{** the default language instead.}%
\else
\language=\csname l@#1\endcsname
\fi
#2}}
\providecommand{\BIBdecl}{\relax}
\BIBdecl

\bibitem{RussellNorvig2020}
S.~J. Russell and P.~Norvig, \emph{Artificial intelligence: A modern approach}.\hskip 1em plus 0.5em minus 0.4em\relax Pearson Education, 2020.

\bibitem{Nilsson1984}
N.~J. Nilsson, ``Shakey the robot,'' SRI International Menlo Park CA Artificial Intelligence Center, Tech. Rep., 1984.

\bibitem{Brooks1991}
R.~A. Brooks, ``Intelligence without representation,'' \emph{Artificial Intelligence}, vol.~47, no. 1-3, pp. 139--159, 1991.

\bibitem{Arkin1998}
R.~C. Arkin, \emph{Behavior-based robotics}.\hskip 1em plus 0.5em minus 0.4em\relax MIT Press, 1998.

\bibitem{Brown2020}
B.~e.~a. Brown, T.\, ``Language models are few-shot learners,'' in \emph{Advances in Neural Information Processing Systems}, vol.~33, 2020, pp. 1877--1901.

\bibitem{Vaswani2017}
A.~Vaswani, N.~Shazeer, N.~Parmar, J.~Uszkoreit, L.~Jones, N.~Gomez, A.\, L.~Kaiser, and I.~Polosukhin, ``Attention is all you need,'' in \emph{Advances in Neural Information Processing Systems}, 2017, pp. 5998--6008.

\bibitem{Laird2012}
J.~E. Laird, \emph{The Soar cognitive architecture}.\hskip 1em plus 0.5em minus 0.4em\relax MIT Press, 2012.

\bibitem{Anderson2004}
J.~R. Anderson, E.~H. Bothell, M.~D. Byrne, S.~Douglass, C.~Lebiere, and Y.~Qin, ``An integrated theory of the mind,'' \emph{Psychological Review}, vol. 111, no.~4, p. 1036, 2004.

\bibitem{Ji2023}
\BIBentryALTinterwordspacing
Z.~Ji, N.~Lee, R.~Frieske, T.~Yu, D.~Su, Y.~Xu, E.~Ishii, Y.~J. Bang, A.~Madotto, and P.~Fung, ``Survey of hallucination in natural language generation,'' \emph{ACM Comput. Surv.}, vol.~55, no.~12, Mar. 2023. [Online]. Available: \url{https://doi.org/10.1145/3571730}
\BIBentrySTDinterwordspacing

\bibitem{Huang2023}
\BIBentryALTinterwordspacing
L.~Huang, W.~Yu, W.~Ma, W.~Zhong, Z.~Feng, H.~Wang, Q.~Chen, W.~Peng, X.~Feng, B.~Qin, and T.~Liu, ``A survey on hallucination in large language models: Principles, taxonomy, challenges, and open questions,'' \emph{ACM Trans. Inf. Syst.}, vol.~43, no.~2, Jan. 2025. [Online]. Available: \url{https://doi.org/10.1145/3703155}
\BIBentrySTDinterwordspacing

\bibitem{shen2023large}
T.~Shen, R.~Jin, Y.~Huang, C.~Liu, W.~Dong, Z.~Guo, X.~Wu, Y.~Liu, and D.~Xiong, ``Large language model alignment: A survey,'' \emph{arXiv preprint arXiv:2309.15025}, 2023.

\bibitem{nasa}
R.~Royce, M.~Kaufmann, J.~Becktor, S.~Moon, K.~Carpenter, K.~Pak, A.~Towler, R.~Thakker, and S.~Khattak, ``Enabling novel mission operations and interactions with rosa: The robot operating system agent,'' NASA/JPL-Caltech Technical Report, 2025.

\bibitem{durante2024agent}
Z.~e.~a. Durante, ``Agent ai: Surveying the horizons of multimodal interaction,'' \emph{arXiv preprint arXiv:2401.03568}, 2024.

\bibitem{McCarthy1959}
J.~McCarthy, ``Programs with common sense,'' in \emph{Mechanisation of Thought Processes}, vol.~1, 1959, pp. 77--84.

\bibitem{FikesNilsson1971}
R.~E. Fikes and N.~J. Nilsson, ``Strips: A new approach to theorem proving applied to problem solving,'' \emph{Artificial Intelligence}, vol.~2, no. 3-4, pp. 189--208, 1971.

\bibitem{McCarthyHayes1969}
J.~McCarthy and P.~J. Hayes, ``Some philosophical problems from the standpoint of artificial intelligence,'' \emph{Machine Intelligence}, vol.~4, pp. 463--502, 1969.

\bibitem{AgreChapman1987}
P.~E. Agre and D.~Chapman, ``Pengi: An implementation of a theory of activity,'' in \emph{AAAI}, 1987, pp. 268--272.

\bibitem{Brooks1986}
R.~A. Brooks, ``A robust layered control system for a mobile robot,'' \emph{IEEE Journal of Robotics and Automation}, vol.~2, no.~1, pp. 14--23, 1986.

\bibitem{PfeiferScheier1999}
R.~Pfeifer and C.~Scheier, \emph{Understanding intelligence}.\hskip 1em plus 0.5em minus 0.4em\relax MIT Press, 1999.

\bibitem{Clark1997}
A.~Clark, \emph{Being there: Putting brain, body, and world together again}.\hskip 1em plus 0.5em minus 0.4em\relax MIT Press, 1997.

\bibitem{Wilson2002}
M.~Wilson, ``Six views of embodied cognition,'' \emph{Psychonomic Bulletin \& Review}, vol.~9, no.~4, pp. 625--636, 2002.

\bibitem{Anderson2007}
J.~R. Anderson, \emph{How can the mind occur in the body?}\hskip 1em plus 0.5em minus 0.4em\relax Oxford University Press, 2007.

\bibitem{BrachmanLevesque1984}
R.~J. Brachman and H.~J. Levesque, ``The tractability of subsumption in frame-based description languages,'' in \emph{AAAI}, 1984, pp. 34--37.

\bibitem{Kumar2024-tz}
P.~Kumar, ``\BIBforeignlanguage{en}{Large language models ({LLMs}): survey, technical frameworks, and future challenges},'' \emph{\BIBforeignlanguage{en}{Artif. Intell. Rev.}}, vol.~57, no.~10, pp. 1--51, Aug. 2024.

\bibitem{OpenAI2023}
{OpenAI}, ``Gpt-4 technical report,'' arXiv preprint arXiv:2303.08774, 2023.

\bibitem{Bubeck2023}
S.~e.~a. Bubeck, ``Sparks of artificial general intelligence: Early experiments with gpt-4,'' arXiv preprint arXiv:2303.12712, 2023.

\bibitem{Mirchev2023}
N.~Mirchev, S.~Jiang, S.~Shah, and A.~Garg, ``Large language models as general-purpose policies for robots,'' arXiv preprint arXiv:2305.05042, 2023.

\bibitem{Ahn2022}
M.~e.~a. Ahn, ``Do as i can, not as i say: Grounding language in robotic affordances,'' arXiv preprint arXiv:2204.01691, 2022.

\bibitem{Brohan2023}
A.~e.~a. Brohan, ``Rt-2: Vision-language-action models transfer web knowledge to robotic control,'' arXiv preprint arXiv:2207.05736, 2023.

\bibitem{Radford2021}
A.~e.~a. Radford, ``Learning transferable visual models from natural language supervision,'' in \emph{International Conference on Machine Learning}.\hskip 1em plus 0.5em minus 0.4em\relax PMLR, 2021, pp. 8748--8763.

\bibitem{Gupta2023}
K.~Gupta, A.\, A.~Yan, J.~Guo, Y.~Cheng, L.~Yang, H.~Chen, L.\, and K.~Keutzer, ``Visual grounding for language-guided navigation,'' in \emph{Proceedings of the IEEE/CVF Winter Conference on Applications of Computer Vision}, 2023, pp. 3562--3572.

\bibitem{Driess2023}
D.~e.~a. Driess, ``Palm-e: An embodiment-aware language model for instruction following with visual and tactile feedback,'' arXiv preprint arXiv:2203.16939, 2023.

\bibitem{Jia2021}
C.~Jia, Y.~T. Chen, Z.~Lu, S.~Tunyasuvunakool, N.~De~Freitas, and D.~Tarlow, ``Scaling up visual and vision-language representation learning with noisy text supervision,'' in \emph{International Conference on Machine Learning}.\hskip 1em plus 0.5em minus 0.4em\relax PMLR, 2021, pp. 4615--4625.

\bibitem{Zhu2023}
K.~Zhu, C.~Gan, L.~Wang, Y.~Fang, X.~Dai, and S.~Han, ``Vision-language models are zero-shot reward function approximators,'' arXiv preprint arXiv:2303.02896, 2023.

\bibitem{Liang2023}
J.~Liang, W.~Huang, F.~Xia, P.~Xu, K.~Hausman, B.~Ichter, P.~Florence, and A.~Zeng, ``Code as policies: Language model-based discrete action policies for embodied ai,'' arXiv preprint arXiv:2209.07753, 2023.

\bibitem{Zeng2023}
A.~e.~a. Zeng, ``Socratic models: Composing zero-shot multimodality with language,'' in \emph{International Conference on Machine Learning}.\hskip 1em plus 0.5em minus 0.4em\relax PMLR, 2023, pp. 26\,944--26\,967.

\bibitem{Yao2023}
S.~Yao, W.~Zhao, J.~Wang, Y.~Cao, S.~Narasimhan, and D.~Zhao, ``React: Synergizing reasoning and acting in language models for task solving,'' arXiv preprint arXiv:2210.03629, 2023.

\bibitem{Cao1997}
Y.~U. Cao, A.~S. Fukunaga, and A.~B. Kahng, ``Cooperative mobile robotics: Antecedents and directions,'' \emph{Autonomous Robots}, vol.~4, no.~1, pp. 7--27, 1997.

\bibitem{Stone2000}
P.~Stone, Ed., \emph{Multiagent systems: A modern approach to distributed artificial intelligence}.\hskip 1em plus 0.5em minus 0.4em\relax MIT Press, 2000.

\bibitem{Wooldridge2009}
M.~Wooldridge, \emph{An introduction to multiagent systems}.\hskip 1em plus 0.5em minus 0.4em\relax John Wiley \& Sons, 2009.

\bibitem{OlfatiSaber2007}
R.~Olfati-Saber, J.~A. Fax, and R.~M. Murray, ``Consensus and cooperation in networked multi-agent systems,'' \emph{Proceedings of the IEEE}, vol.~95, no.~1, pp. 215--233, 2007.

\bibitem{Bradshaw2017}
J.~M. Bradshaw, R.~R. Hoffman, M.~Johnson, and P.~J. Feltovich, ``Beyond human-centered autonomy: Collaboration and reciprocal adaptation in human-agent teams,'' \emph{IEEE Intelligent Systems}, vol.~32, no.~3, pp. 70--78, 2017.

\bibitem{Longo2023}
M.~Longo, L.~Rathenau, and J.~Weber, \emph{Ethical autonomy in intelligent agents: Embedding ethics into artificial intelligence}.\hskip 1em plus 0.5em minus 0.4em\relax Springer Nature, 2023.

\bibitem{BeerRandallFitch2014}
R.~D. Beer, M.~Randall, and P.~Fitch, ``Evolving dynamical neural networks for adaptive behavior,'' \emph{Adaptive Behavior}, vol.~22, no.~1, pp. 3--27, 2014.

\bibitem{Maynez2020}
J.~Maynez, S.~Narayan, L.~Lokhande, and R.~Reddy, ``On faithfulness and hallucination in abstractive summarization,'' in \emph{Proceedings of the 58th Annual Meeting of the Association for Computational Linguistics}, 2020, pp. 1883--1896.

\bibitem{VerhelstMoons2017}
M.~Verhelst and E.~Moons, ``What edge computing can do for deep learning,'' in \emph{2017 IEEE International Conference on Computer Communications Workshops (INFOCOM WKSHPS)}.\hskip 1em plus 0.5em minus 0.4em\relax IEEE, 2017, pp. 1--6.

\bibitem{Amodei2016}
D.~Amodei, C.~Olah, J.~Steinhardt, P.~Christiano, J.~Schulman, and I.~Sutskever, ``Concrete ai safety problems,'' arXiv preprint arXiv:1606.06565, 2016.

\bibitem{KoopmanWagner2017}
P.~Koopman and M.~Wagner, ``Challenges in autonomous vehicle verification and validation,'' \emph{SAE International Journal of Transportation Safety}, vol.~5, no.~1, pp. 19--27, 2017.

\bibitem{Lin2011}
P.~Lin, K.~Abney, and R.~Jenkins, ``Robot ethics: Mapping the issues for computer scientists,'' \emph{AI Magazine}, vol.~32, no.~1, p.~15, 2011.

\bibitem{WallachAllen2008}
W.~Wallach and C.~Allen, \emph{Moral machines: Teaching robots right from wrong}.\hskip 1em plus 0.5em minus 0.4em\relax Oxford University Press, 2008.

\bibitem{Wei2022CoT}
J.~Wei, X.~Wang, D.~Schuurmans, M.~Bosma, B.~Ichter, F.~Xia, E.~Chi, Q.~Le, and D.~Zhou, ``Chain-of-thought prompting elicits reasoning in large language models,'' arXiv preprint arXiv:2201.11903, 2022.

\bibitem{Madaan2023SelfRefine}
A.~e.~a. Madaan, ``Self-refine: Iterative refinement with self-feedback,'' arXiv preprint arXiv:2303.17651, 2023.

\bibitem{willners2021market}
J.~S. Willners, I.~Carlucho, T.~Łuczyński, S.~Katagiri, C.~Lemoine, J.~Roe, D.~Stephens, S.~Xu, Y.~Carreno, Èric Pairet, C.~Barbalata, Y.~Petillot, and S.~Wang, ``From market-ready rovs to low-cost auvs,'' in \emph{OCEANS 2021: San Diego--Porto}.\hskip 1em plus 0.5em minus 0.4em\relax IEEE, 2021, pp. 1--7.

\bibitem{buchholz2025framework}
M.~Buchholz, I.~Carlucho, Z.~Huang, M.~Grimaldi, P.~Nicolay, S.~Tunçay, and Y.~R. Petillot, ``Framework for robust motion planning of tethered multi-robot systems in marine environments,'' in \emph{Proceedings of the IEEE/MTS OCEANS Conference}.\hskip 1em plus 0.5em minus 0.4em\relax Brest, France: IEEE, May 2025.

\bibitem{buchholz2025tethered}
M.~Buchholz, I.~Carlucho, M.~Grimaldi, and Y.~R. Petillot, ``Tethered multi-robot systems in marine environments,'' in \emph{Proceedings of the ICRA 2025 AQU2ASIM Workshop on Marine Robotics}, Atlanta, USA, 2025.

\bibitem{buchholz2025contextaware}
\BIBentryALTinterwordspacing
M.~Buchholz, I.~Carlucho, M.~Grimaldi, M.~Koskinopoulou, and Y.~R. Petillot, ``Context-aware behavior learning with heuristic motion memory for underwater manipulation,'' 2025. [Online]. Available: \url{https://arxiv.org/abs/2507.14099}
\BIBentrySTDinterwordspacing

\bibitem{morgan2022autonomous}
E.~Morgan, I.~Carlucho, W.~Ard, and C.~Barbalata, ``Autonomous underwater manipulation: Current trends in dynamics, control, planning, perception, and future directions,'' \emph{Current Robotics Reports}, vol.~3, no.~4, pp. 187--198, 2022.

\bibitem{ardupilot_software}
{The ArduPilot Development Team}, ``{ArduPilot}: Open source autopilot software suite,'' \url{https://ardupilot.org}, 2025, accessed: 5-July-2025.

\bibitem{8867434}
P.~Cieślak, ``Stonefish: An advanced open-source simulation tool designed for marine robotics, with a ros interface,'' in \emph{OCEANS 2019 - Marseille}, 2019, pp. 1--6.

\bibitem{grimaldi2025stone}
\BIBentryALTinterwordspacing
M.~Grimaldi, P.~Cieslak, E.~Ochoa, V.~Bharti, H.~Rajani, I.~Carlucho, M.~Koskinopoulou, Y.~R. Petillot, and N.~Gracias, ``Stonefish: Supporting machine learning research in marine robotics,'' 2025. [Online]. Available: \url{https://arxiv.org/abs/2502.11887}
\BIBentrySTDinterwordspacing

\end{thebibliography}
\end{document}